%% file: main.tex
\documentclass[10pt,twocolumn,letterpaper]{article}

\usepackage[pagenumbers]{cvpr} 

\newcommand{\ours}{MagicAnimate}
\newcommand{\projpage}{https://showlab.github.io/magicanimate}

\input{preamble}

\definecolor{cvprblue}{rgb}{0.21,0.49,0.74}
\usepackage[pagebackref,breaklinks,colorlinks,citecolor=cvprblue]{hyperref}
\usepackage{multirow}
\usepackage[dvipsnames]{xcolor}
\usepackage{pifont}
\newcommand{\cmark}{\ding{51}}
\newcommand{\xmark}{\ding{55}}
\def\paperID{5454} 
\def\confName{CVPR}
\def\confYear{2024}

\title{\ours{}: Temporally Consistent Human Image \\Animation using Diffusion Model}

\author{Zhongcong Xu\(^1\) \quad Jianfeng Zhang\(^2\) \quad Jun Hao Liew\(^2\) \quad Hanshu Yan\(^2\) \quad Jia-Wei Liu\(^1\)\\
Chenxu Zhang\(^2\) \quad Jiashi Feng\(^2\) \quad Mike Zheng Shou\(^1\)\thanks{Corresponding author}\\
\(^1\)Show Lab, National University of Singapore\quad \(^2\)ByteDance\\
{\tt\small zhongcongxu@u.nus.edu}
}

\begin{document}
\maketitle
\input{sec/0_abstract}    
\input{sec/1_introduction}
\input{sec/2_related_works}
\input{sec/3_method}
\input{sec/4_experiments}

\input{sec/6_conclusion}
\input{sec/7_acknowledgement}
{
    \small
    \bibliographystyle{ieeenat_fullname}
    \bibliography{main}
}

\end{document}

%% file: preamble.tex
\usepackage[dvipsnames]{xcolor}

\DeclareMathAlphabet{\mathbsf}{OT1}{cmss}{bx}{n}
\DeclareMathAlphabet{\mathssf}{OT1}{cmss}{m}{sl}

\DeclareSymbolFont{bsfletters}{OT1}{cmss}{bx}{n}  
\DeclareSymbolFont{ssfletters}{OT1}{cmss}{m}{n}
\DeclareMathSymbol{\bsfGamma}{0}{bsfletters}{'000}
\DeclareMathSymbol{\ssfGamma}{0}{ssfletters}{'000}
\DeclareMathSymbol{\bsfDelta}{0}{bsfletters}{'001}
\DeclareMathSymbol{\ssfDelta}{0}{ssfletters}{'001}
\DeclareMathSymbol{\bsfTheta}{0}{bsfletters}{'002}
\DeclareMathSymbol{\ssfTheta}{0}{ssfletters}{'002}
\DeclareMathSymbol{\bsfLambda}{0}{bsfletters}{'003}
\DeclareMathSymbol{\ssfLambda}{0}{ssfletters}{'003}
\DeclareMathSymbol{\bsfXi}{0}{bsfletters}{'004}
\DeclareMathSymbol{\ssfXi}{0}{ssfletters}{'004}
\DeclareMathSymbol{\bsfPi}{0}{bsfletters}{'005}
\DeclareMathSymbol{\ssfPi}{0}{ssfletters}{'005}
\DeclareMathSymbol{\bsfSigma}{0}{bsfletters}{'006}
\DeclareMathSymbol{\ssfSigma}{0}{ssfletters}{'006}
\DeclareMathSymbol{\bsfUpsilon}{0}{bsfletters}{'007}
\DeclareMathSymbol{\ssfUpsilon}{0}{ssfletters}{'007}
\DeclareMathSymbol{\bsfPhi}{0}{bsfletters}{'010}
\DeclareMathSymbol{\ssfPhi}{0}{ssfletters}{'010}
\DeclareMathSymbol{\bsfPsi}{0}{bsfletters}{'011}
\DeclareMathSymbol{\ssfPsi}{0}{ssfletters}{'011}
\DeclareMathSymbol{\bsfOmega}{0}{bsfletters}{'012}
\DeclareMathSymbol{\ssfOmega}{0}{ssfletters}{'012}

\usepackage{thmtools, thm-restate}



%% file: sec/0_abstract.tex
\begin{abstract}
This paper studies the human image animation task, which aims to generate a video of a certain reference identity following a particular motion sequence. Existing animation works typically employ the frame-warping technique to animate the reference image towards the target motion. Despite achieving reasonable results, these approaches face challenges in maintaining temporal consistency throughout the animation due to the lack of temporal modeling and poor preservation of reference identity. In this work, we introduce \ours{}, a diffusion-based framework that aims at enhancing temporal consistency, preserving reference image faithfully, and improving animation fidelity. To achieve this, we first develop a video diffusion model to encode temporal information. Second, to maintain the appearance coherence across frames, we introduce a novel appearance encoder to retain the intricate details of the reference image. Leveraging these two innovations, we further employ a simple video fusion technique to encourage smooth transitions for long video animation. Empirical results demonstrate the superiority of our method over baseline approaches on two benchmarks. Notably, our approach outperforms the strongest baseline by over 38\% in terms of video fidelity on the challenging TikTok dancing dataset. Code and model will be made available. 
\end{abstract}

%% file: sec/1_introduction.tex
\section{Introduction}
\label{sec:introduction}
\begin{figure}[ht]
\centering
\includegraphics[width=0.9\linewidth]{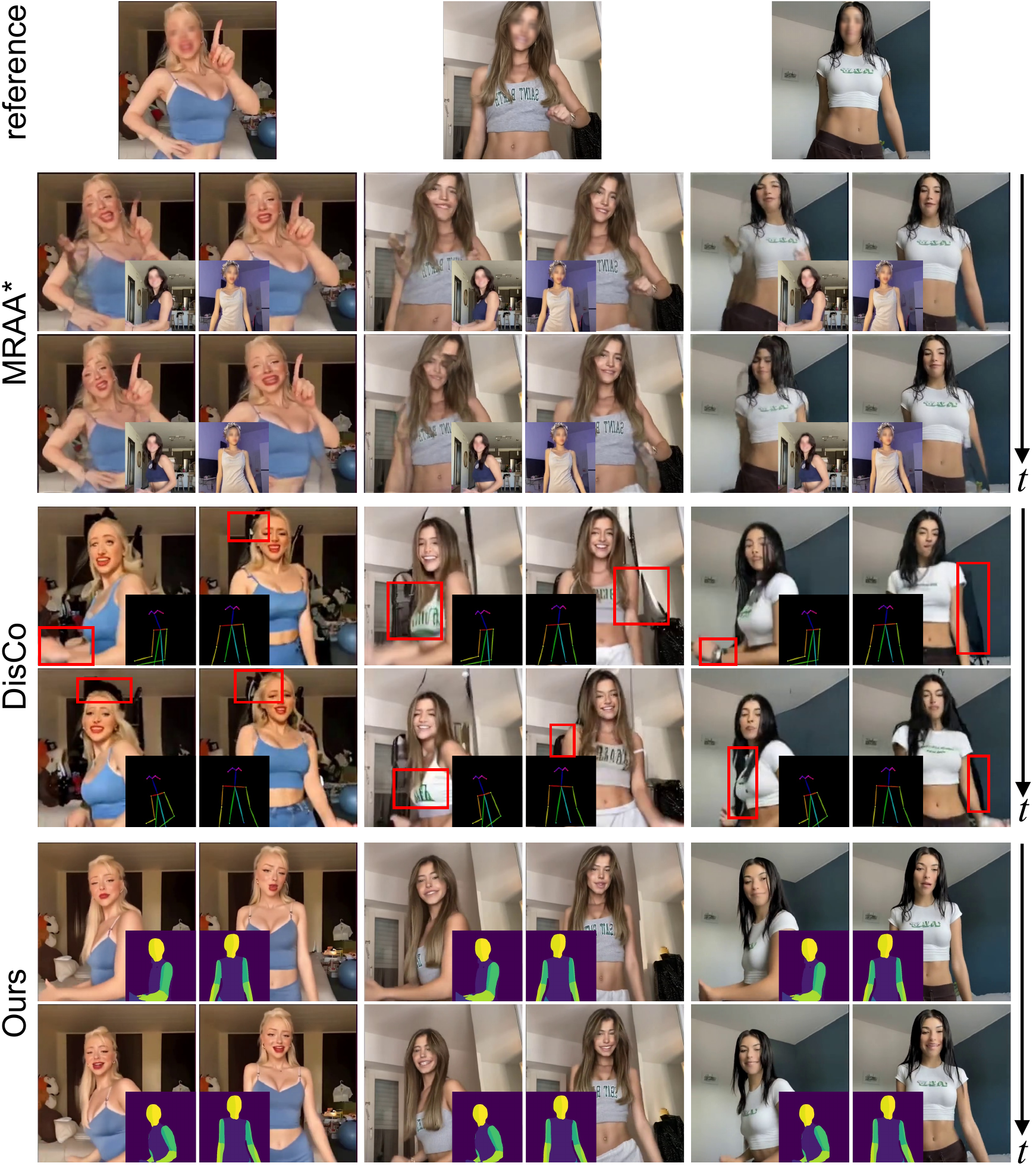}
\vspace{-0.5em}
\caption{
Given a sequence of motion signals, \ours{} produces temporally consistent animation for reference identity images, whereas state-of-the-art methods fail to generalize or preserve the reference appearance, as highlighted in \textcolor{red}{red} boxes. 
The motion sequence is overlaid at the corner. \(^*\)Note that MRAA directly uses video frames as the driving signal. The complete video results can be found on our \href{\projpage}{Project Page}.}
\label{fig:app1:crossid}
\vspace{-1.2em}
\end{figure}
Given a sequence of motion signals such as video, depth, or pose, the image animation task aims to bring static images to life. The animation of humans, animals, cartoons, or other general objects, has attracted much attention in research~\cite{siarohin2019first,siarohin2021motion,zhao2022thin}. Among these, human image animation~\cite{wang2023disco,karras2023dreampose,zhang2023magicavatar} has been the most extensively explored, 
given its potential applications across various domains, including social media, movie industry, and entertainment, \etc In contrast to traditional graphic approaches~\cite{xiang2021modeling,guo2019relightables}, the abundance of data enables the development of low-cost data-driven animation frameworks~\cite{wang2021one,geng20193d,chan2019everybody,xu2023xagen,zhang2023avatargen,hong2023evad}.

Existing data-driven methods for human image animation can be categorized into two primary groups based on the generative backbone models used, namely GAN-based and diffusion-based frameworks. The former~\cite{siarohin2019first,wang2021one} typically employs a warping function to deform the reference image into the target pose and utilize GAN models to extrapolate the missing or occluded body parts. In contrast, the latter~\cite{wang2023disco,karras2023dreampose} harness appearance~\cite{radford2021learning} and pose conditions~\cite{zhang2023adding} to generate the target image based on pretrained diffusion models~\cite{rombach2022high}.
Despite generating visually plausible animations, these methods typically exhibit several limitations: 1) GAN-based methods possess restricted motion transfer capability, resulting in unrealistic details in occluded regions and limited generalization ability for cross-identity scenarios, as depicted in Figure~\ref{fig:app1:crossid}. 2) Diffusion-based methods, on the other hand, process a lengthy video in a frame-by-frame manner and then stack results along the temporal dimension. Such approaches neglect temporal consistency, resulting in flickering results. 
In addition, these works typically rely on CLIP~\cite{radford2021learning} to encode reference appearance, which is known to be less effective in preserving details, as highlighted in the red boxes in Figure~\ref{fig:app1:crossid}.

In this work, to address the aforementioned limitations, we develop a human image animation framework called \ours{} that offers long-range temporal consistency, robust appearance encoding, and high per-frame quality. To achieve this, we first develop a video diffusion model that encodes temporal information by incorporating temporal attention blocks into the diffusion network. Secondly, we introduce an innovative appearance encoder to preserve the human identity and background information derived from the reference image. Unlike existing works that employ CLIP-encoded visual features, our appearance encoder is capable of extracting 
dense visual features to guide the animation, which leads to better preservation of identity, background, clothes, \etc.
To further improve per-frame fidelity, we additionally devise an image-video joint training strategy to leverage diverse single-frame image data for augmentation, which provides richer visual cues to improve the modeling capability of our framework for details.
Lastly, we leverage a surprisingly simple video fusion technique to enable long video animation with smooth transitions. 

In summary, our contributions are three-fold: 
(1) We propose \ours{}, a novel diffusion-based human image animation approach that integrates temporal consistency modeling, precise appearance encoding, and temporal video fusion, for synthesizing temporally consistent human animation of arbitrary length. 
(2) Our method achieves state-of-the-art performance on two benchmarks. Notably, it surpasses the strongest baseline by more than 38\% in terms of video quality on the challenging TikTok dancing dataset. 
(3) \ours{} showcases robust generalization ability, supporting cross-identity animation and various downstream applications, including unseen domain animation and multi-person animation.

%% file: sec/2_related_works.tex
\section{Related Work}
\label{sec:related_work}
\subsection{Data-driven Animation}
Prior efforts in image animation have predominantly concentrated on the human body or face, leveraging the abundance of diverse training data and domain-specific knowledge, such as keypoints~\cite{cao2014displaced,zakharov2019few,qian2019make}, semantic parsing~\cite{nirkin2019fsgan}, and statistical parametric models~\cite{xu2023omniavatar,xu2023xagen,zhang2023avatargen,thies2016face2face}. Building upon these motion signals, a long line of work~\cite{thies2016face2face,wang2022latent,siarohin2019animating,zakharov2020fast,wang2021one,siarohin2023unsupervised} has emerged. These approaches can be classified into two categories based on their animation pipeline, \ie, implicit and explicit animation. Implicit animation methods transform the source image to the target motion signal by deforming the reference image in sub-expression space~\cite{thies2016face2face} or manipulating the latent space of a generative model~\cite{wang2022latent,oorloff2023robust,tulyakov2018mocogan,qian2019make}. The generative backbone conditions on target motion signal to synthesize animations. Conversely, explicit methods warp the source image to the target by either 2D optical flow~\cite{siarohin2019animating,siarohin2019first,siarohin2021motion,zhao2022thin,ren2021pirenderer,zakharov2020fast}, 3D deformation field~\cite{wang2021one,mallya2022implicit,cao2014displaced}, or directly sawpping the face of target image~\cite{nirkin2019fsgan}. 
In addition to deforming the source image or 3D mesh, recent research efforts~\cite{siarohin2023unsupervised,xu2023omniavatar,zhang2023avatargen,xu2023xagen} explore explicitly deforming points in 3D neural representations for human body and face synthesis, showcasing improved temporal and multi-view consistency.
\subsection{Diffusion Models for Animation}
The remarkable progress in diffusion models~\cite{rombach2022high,song2020denoising,saharia2022photorealistic} has propelled text-to-image generation to unprecedented success, spawning numerous subsequent works, such as controllable image generation~\cite{zhang2023adding} and video generation~\cite{wu2023tune}, \etc. Recent works have embraced diffusion models for human-centric video generation~\cite{ma2023follow} and animation~\cite{wang2023disco}. Among these works, a common approach~\cite{wang2023leo} develops a diffusion model for generating 2D optical flow and then animates the reference image using frame-warping technique~\cite{siarohin2019first}.
Moreover, many diffusion-based animation frameworks~\cite{wang2023disco, karras2023dreampose,zhang2023magicavatar} employ Stable Diffusion~\cite{rombach2022high} as their image generation backbone and leverage ControlNet~\cite{zhang2023adding} to condition the animation process on OpenPose~\cite{cao2017realtime} keypoint sequences. For the reference image condition, they usually adopt a pretrained image-language model, CLIP~\cite{radford2021learning}, to encode the image into a semantic-level text token space and guide the image generation process through cross-attention. While these works yield visually plausible results, most of them process each video frame independently and neglect the temporal information in animation videos, which inevitably leads to flickering animation results. 

%% file: sec/3_method.tex
\begin{figure*}[t]
\centering
\includegraphics[width=0.83\textwidth]{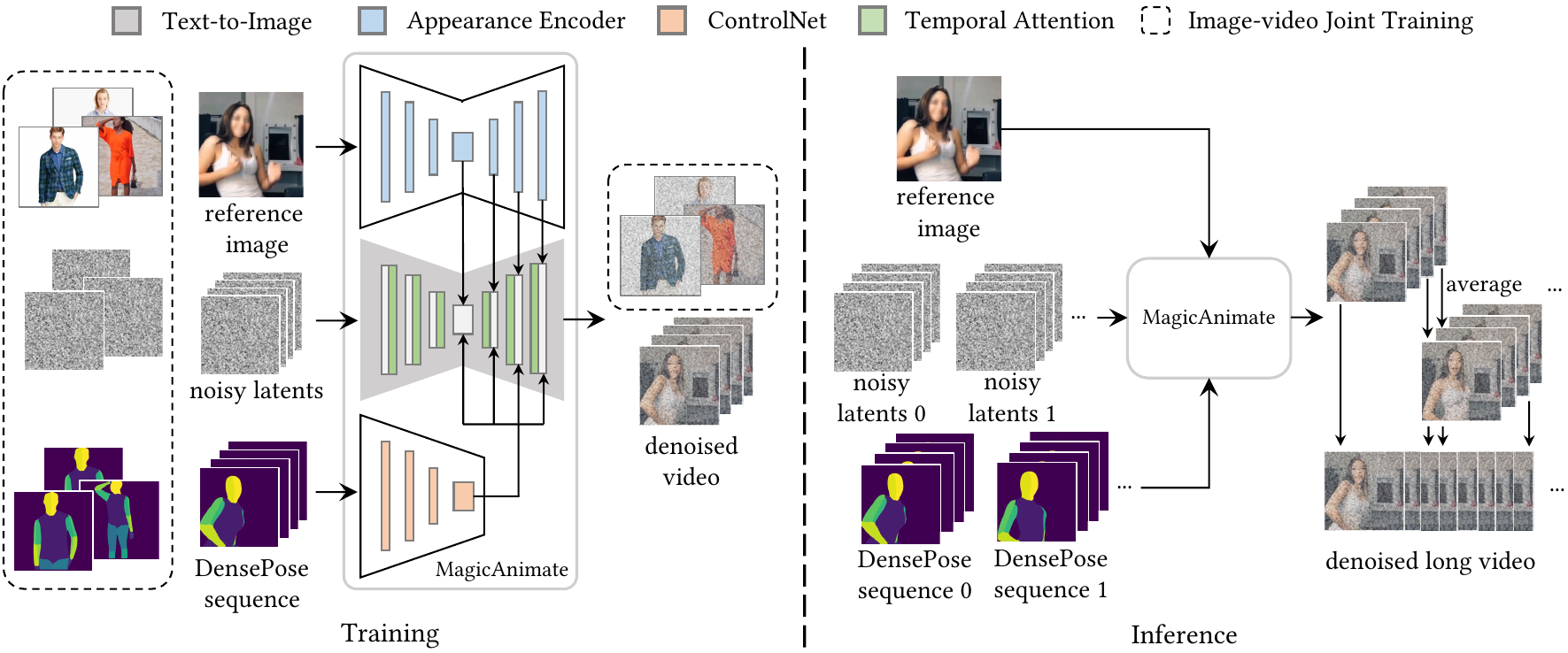}
\vspace{-0.9em}
\caption{
\textbf{\ours{} pipeline}. Given a reference image and the target DensePose motion sequence, \ours{} employs a video diffusion model and an appearance encoder for temporal modeling and identity preserving, respectively (\textbf{left panel}). 
To support long video animation, we devise a simple video fusion strategy that produces smooth video transition during inference (\textbf{right panel}).}
\label{fig:framework}
\vspace{-1.5em}
\end{figure*}
\section{Method}
\label{method}
Given a reference image \(I_{\text{ref}}\) and a motion sequence \(\boldsymbol{p}^{1:N} = [\boldsymbol{p}_1, \cdots, \boldsymbol{p}_N]\), where $N$ is the number of frames, our objective is to synthesize a continuous video \(I^{1:N}=[I_1, \cdots, I_N]\) with the appearance of \(I_{\text{ref}}\) while adhering to the provided motion \(\boldsymbol{p}^{1:N}\).

Existing diffusion-based frameworks~\cite{wang2023disco,karras2023dreampose} process each frame independently,
neglecting the temporal consistency among different frames, which consequently results in flickering animations.
To address this, we build a video diffusion model \(\mathcal{F}^{\text T}\) for temporal modeling by incorporating temporal attention blocks into the diffusion backbone (Sec.~\ref{temp}).
In addition, existing works~\cite{wang2023disco,karras2023dreampose} use CLIP~\cite{radford2021learning} encoder to encode the reference image. We argue that these semantic-level features are too sparse and compact to capture intricate details. Therefore, we introduce a novel appearance encoder \(\mathcal{F}_{\text a}\) (Sec.~\ref{app_enc}) to encode \(I_{\text{ref}}\) into appearance embedding \(\boldsymbol{y}_{\text a}\) and condition our model on it for identity- and background-preserving animation.

The overall pipeline of our \ours{} (Sec.~\ref{pipe}) is depicted in Figure~\ref{fig:framework}.
We first embed the reference image into appearance embedding \(\boldsymbol{y}_{\text a}\) using our appearance encoder. We then pass the target pose sequence, \ie, DensePose~\cite{guler2018densepose}, into a pose ControlNet~\cite{zhang2023adding} \(\mathcal{F}_{\text p}\) to extract motion condition \(\boldsymbol{y}_{\text p}^{1:K}\). 
Conditioning on these two signals, our video diffusion model is trained to animate the reference human identity to follow the given motions. In practice, due to memory constraints, we process the entire video in a segment-by-segment manner. Thanks to the temporal modeling and robust appearance encoding, \ours{} can largely maintain temporal and appearance consistency across segments. Nevertheless, there still exists minor discontinuities between segments. To mitigate this, we leverage a simple video fusion approach to improve the transition smoothness. Specifically, as depicted in Figure~\ref{fig:framework}, we decompose the entire video into overlapping segments and simply average the predictions for overlapping frames. 
Lastly, we also introduce an image-video joint training strategy to further enhance the reference-preserving capability and single-frame fidelity (Sec.~\ref{joint}).

\subsection{Temporal Consistency Modeling}
\label{temp}
To ensure temporal consistency across video frames, we extend the image diffusion model to the video domain. Specifically, we inflate the original 2D UNet to 3D temporal UNet by inserting temporal attention layers~\cite{wu2023tune,zhou2022magicvideo,guo2023animatediff}. The temporal UNet is denoted as $\mathcal{F}^{\text T}(\cdot; \theta^{\text T})$ with trainable parameters $\theta^{\text T}$. The architecture of the inflated UNet blocks is illustrated in Figure~\ref{fig:framework}. 
We begin with randomly initialized latent noise \(\boldsymbol{z}_{\boldsymbol{t}}^{1:K}\) where \(K\) is the length of the video frames. 
We then stack \(K\) consecutive poses into a DensePose sequence \(\boldsymbol{p}^{1:K}\) for motion guidance. We input \(\boldsymbol{z}_{\boldsymbol{t}}^{1:K}\) to our video diffusion backbone \(\mathcal{F}^{\text T}\) by reshaping the input features from \(\mathbb{R}^{N\times C \times K \times H \times W}\) into \(\mathbb{R}^{(NK)\times C \times H \times W}\). Within temporal modules, we reshape the features into \(\mathbb{R}^{(NHW)\times K\times C}\) to compute cross-frame information along the temporal dimension.  Following prior works~\cite{guo2023animatediff}, we add sinusoidal positional encoding to make the model aware of the position for each frame within the video. As such, we compute temporal attention using the standard attention operation, which is formulated as \(\text{Attention}(Q, K, V)=\text{Softmax}(\frac{QK^T}{\sqrt{d}})V\), where \(Q=W^Q\boldsymbol{z}_{\boldsymbol{t}}\), \(K=W^K\boldsymbol{z}_{\boldsymbol{t}}\), \(V=W^V\boldsymbol{z}_{\boldsymbol{t}}\). Through this attention mechanism, \ours{} aggregates temporal information from neighboring frames and synthesizes \(K\) frames with improved temporal consistency.

\subsection{Appearance Encoder}
\label{app_enc}
The goal of human image animation is to generate results under the guidance of a reference image \(I_{\text{ref}}\). 
The core objective of our appearance encoder is representing \(I_{\text{ref}}\) with detailed identity- and background-related features that can be injected into our video diffusion model for retargeting under the motion signal guidance.
Inspired by recent works on dense reference image conditioning, such as MasaCtrl~\cite{cao2023masactrl} and Reference-only ControlNet~\cite{refonly}, we propose a novel appearance encoder with improved identity and background preservation to enhance single-frame fidelity and temporal coherence. Specifically, our appearance encoder 
creates another trainable copy of the base UNet \(\mathcal{F}_{\text a}(\cdot; \theta^{\text a})\) and compute the condition features for the reference image \(I_\text{ref}\) for each denoising step \(\boldsymbol{t}\). This process is mathematically formulated as
\begin{equation}
\boldsymbol{y}_{\text a}=\mathcal{F}_{\text a}(\boldsymbol{z}_{\boldsymbol{t}}|I_\text{ref},\boldsymbol{t}, \theta^{\text a}),
\end{equation}
where \(\boldsymbol{y}_{\text a}\) is a set of normalized attention hidden states for the middle and upsampling blocks. Different from ControlNet which adds conditions in a residual manner, \textit{we pass these features to the spatial self-attention layers in the UNet blocks} by concatenating each feature in \(\boldsymbol{y}_{\text a}\) with the original UNet self-attention hidden states to inject the appearance information. 
Our appearance condition process is mathematically formulated as:
\vspace{-1em}
\begin{equation}
\begin{aligned}
    &\text{Attention}(Q, K, V,\boldsymbol{y}_{\text a})=\text{Softmax}(\frac{QK'^T}{\sqrt{d}})V', \\
    Q&=W^Q\boldsymbol{z}_{\boldsymbol{t}}, K' = W^K[\boldsymbol{z}_{\boldsymbol{t}},\boldsymbol{y}_{\text a}], V' = W^V[\boldsymbol{z}_{\boldsymbol{t}},\boldsymbol{y}_{\text a}],
\end{aligned}
\end{equation} 
where \([\cdot]\) denotes concatenation operation. Through this operation, we can adapt the spatial self-attention mechanism in our video diffusion model into a hybrid one. This hybrid attention mechanism can not only maintain the semantic layout of the synthesized image, such as the pose and position of the human in the image, but also query the contents from the reference image in the denoising process to preserve the details, including identity, clothes, accessories, and background. This improved preservation capability benefits our framework in two aspects: (1) our method can transfer the reference image faithfully to the target motion; (2) the strong appearance condition contributes to temporal consistency by retaining the same identity, background, and other details throughout the entire video.

\subsection{Animation Pipeline}
\label{pipe}
With the incorporation of temporal consistency modeling and the appearance encoder, we combine these elements with pose conditioning, \ie, ControlNet~\cite{zhang2023adding}, to transform the reference image to the target poses.

\noindent{\bf Motion transfer.}
ControlNet for OpenPose~\cite{cao2017realtime} keypoints is commonly employed for animating reference human images. Although it produces reasonable results, we argue that the major body keypoints are sparse and not robust to certain motions, such as rotation.  
Consequently, we choose DensePose~\cite{guler2018densepose} as the motion signal \(\boldsymbol{p}_i\) for dense and robust pose conditions. We employ a pose ControlNet \(\mathcal{F}_{\text p}(\cdot, \theta^{\text p})\), the pose condition for frame $i$ is computed as
\begin{equation}
\boldsymbol{y}_{{\text p},i}=\mathcal{F}_{\text p}(\boldsymbol{z}_{\boldsymbol{t}}|\boldsymbol{p}_i,\boldsymbol{t}, \theta^{\text p}),
\end{equation}
where \(\boldsymbol{y}_{{\text p},i}\) is a set of condition residuals added to the residuals for the middle and upsampling blocks in the diffusion model. In our pipeline, we concatenate the motion feature of each pose in a DensePose sequence into \(\boldsymbol{y}_{\text p}^{1:K}\).

\noindent {\bf Denoising process.} Building upon the appearance condition \(\boldsymbol{y}_{\text a}\) and motion condition \(\boldsymbol{y}_{\text p}^{1:K}\), \ours{} animates the reference image following the DensePose sequence. The noise estimation function \(\epsilon^{1:K}_{\theta}(\cdot)\) in the denoising process is mathematically formulated as:
\begin{equation}
\epsilon^{1:K}_\theta\left(\boldsymbol{z}^{1:K}_t, \boldsymbol{t}, I_\text{ref}, \boldsymbol{p}^{1:K} \right)=\mathcal{F}^{\text T}(\boldsymbol{z}^{1:K}_{\boldsymbol{t}}|\boldsymbol{t}, \boldsymbol{y}_{\text a},  \boldsymbol{y}^{1:K}_p),
\end{equation}
where $\theta$ is the collection of all the trainable parameters, namely $\theta^{\text T}$, $\theta^{\text a}$, and $\theta^{\text p}$.

\smallskip
\noindent {\bf Long video animation.}
With temporal consistency modeling and appearance encoder, we can generate temporally consistent human image animation results for arbitrary length via segment-by-segment processing.
However, unnatural transitions and inconsistent details across segments may occur because temporal attention blocks cannot model long-range consistency across different segments.

To address this challenge, we employ a sliding window method to improve transition smoothness in the inference stage. As shown in Figure~\ref{fig:framework}, we divide the long motion sequence into multiple segments with temporal overlap, where each segment has a length of \(K\). We first sample noise \(\boldsymbol{z}^{1:N}\) for the entire video with \(N\) frames, and also partition it into noise segments with overlap \(\{\boldsymbol{z}^{1:K}, \boldsymbol{z}^{K-s+1:2K-s}, ..., \boldsymbol{z}^{n(K-s)+1:n(K-s)+K}\}\), where \(n=\lceil(N-K)/(K-s)\rceil\) and \(s\) is the overlap stride, with \(s<K\). 
If \(\left(N-K\right) \mod{\left(K-s\right)} \neq 0\), \ie, the last segment size is less than $K$, for simplicity, we simply pad it with the first few frames to construct a $K$-frame segment.
Besides, we empirically find that sharing the same initial noise \(\boldsymbol{z}^{1:K}\) for all the segments improves video quality. For each denoising timestep \(\boldsymbol{t}\), we predict noise and obtain \(\epsilon_{\theta}^{1:K}\) for each segment, and then merge them into \(\epsilon_{\theta}^{1:N}\) by averaging overlap frames. When \(\boldsymbol{t}=0\), we obtain the final animation video \(I^{1:N}\).

\subsection{Training}
\label{joint}

\noindent {\bf Learning objectives.} 
We employ a multi-stage training strategy for our \ours{}. In the first stage, we omit the temporal attention layers temporarily and train the appearance encoder together with pose ControlNet. The loss term of this stage is computed as
\begin{equation}
    \mathcal{L}_1=\mathbb{E}_{\boldsymbol{z}_0, \boldsymbol{t}, I_\text{ref},\boldsymbol{p}_i, \epsilon \sim \mathcal{N}(0,1)}\left[\| \epsilon-\epsilon_\theta \|_2^2\right],
\end{equation}
where 
$\boldsymbol{p}_i$ is the DensePose of target image \(I_i\). The learnable modules are \(\mathcal{F}_{\text p}(\cdot, \theta^{\text p})\) and \(\mathcal{F}_{\text a}(\cdot, \theta^{\text a})\). In the second stage, we optimize only the temporal attention layers in \(\mathcal{F}^{\text T}(\cdot, \theta^{\text T})\), and the learning objective is formulated as
\begin{equation}
   \mathcal{L}_2=\mathbb{E}_{\boldsymbol{z}^{1:K}_0, \boldsymbol{t}, I_\text{ref},\boldsymbol{p}^{1:K}, \epsilon^{1:K} \sim \mathcal{N}(0,1)}\left[\| \epsilon^{1:K}-\epsilon^{1:K}_\theta\|_2^2\right].
\end{equation}
\noindent{\bf Image-video joint training.} 
Human video datasets~\cite{siarohin2021motion,jafarian2021learning}, compared with image datasets, have a much smaller scale and are less diverse in terms of identities, backgrounds, and poses. This restricts the effective learning of reference condition capability of our animation framework. To alleviate this issue, we employ an image-video joint training strategy. 

In the first stage when we pretrain the appearance encoder and pose ControlNet, we set a probability threshold \(\tau_0\) for sampling the human images from a large-scale image dataset~\cite{schuhmann2021laion}. We draw a random number \(r \sim U(0,1)\), where \(U\left(\cdot, \cdot\right)\) denotes uniform distribution. If \(r\leq\tau_0\), we use the sampled image for training. In this case, the conditioning pose \(\boldsymbol{p}_i\) is estimated from \(I_\text{ref}\), and the learning objective of our framework becomes reconstruction.

Although the introduction of temporal attention in the second stage helps improve temporal modeling, we also notice that this leads to degraded per-frame quality.

To simultaneously improve temporal coherence and maintain single-frame image fidelity, we also employ joint training in this stage. Specifically, we select two probability thresholds \(\tau_1\) and \(\tau_2\) empirically, and compare \(r \sim U(0,1)\) with these thresholds. When \(r \leq \tau_1\), we sample the training data from the image dataset, and we sample data from the video dataset otherwise. 

Based on the different training data, our denoising process in the training stage is formulated as
\begin{equation}
    \epsilon_\theta^{1:K} = \begin{cases}
    \epsilon_\theta^{1:K}\left(\boldsymbol{z}_t, \boldsymbol{t}, I_\text{ref}, \boldsymbol{p}_i \right), \text{with}~i = \text{ref}, &\text{if}~ r \leq \tau_1,\\
    \epsilon_\theta^{1:K}\left(\boldsymbol{z}_t, \boldsymbol{t}, I_\text{ref}, \boldsymbol{p}_i \right), \text{with}~i \neq \text{ref}, &\text{if}~ \tau_1 \leq r \leq \tau_2.\\
    \epsilon_\theta^{1:K}\left(\boldsymbol{z}^{1:K}_t, \boldsymbol{t}, I_\text{ref}, \boldsymbol{p}^{1:K} \right), &\text{if}~ r \geq \tau_2
    \end{cases}
\end{equation}

%% file: sec/4_experiments.tex
\section{Experiments}
\label{experiments}
We evaluate the performance of \ours{} using two datasets, namely TikTok~\cite{jafarian2021learning} and TED-talks~\cite{siarohin2021motion}. The TikTok dataset comprises 350 dancing videos, while TED-talks includes 1,203 video clips extracted from TED-talk videos on YouTube. To ensure a fair comparison with state-of-the-art methods, we utilize the identical test set as DisCo~\cite{wang2023disco} for TikTok evaluation and adhere to the official train/test split for TED-talks.
All datasets undergo the same preprocessing pipeline. Please refer to {\it Sup. Mat.} for our dataset preprocessing and implementation details. 

\begin{table*}[t]
 \small
  \centering
    \begin{subtable}[t]{\linewidth}
    \centering
  \begin{tabular}{lccccccc}
    \toprule
    \multirow{2}{*}{Method~~}&\multicolumn{5}{c}{Image} &\multicolumn{2}{c}{Video}\\
    \cmidrule(r){2-6} \cmidrule(r){7-8} & L1\(\downarrow\)~~ & PSNR\(\uparrow\)~~  &SSIM\(\uparrow\)~~ &LPIPS\(\downarrow\)~~ & FID\(\downarrow\)~~ & FID-VID\(\downarrow\)~~& FVD\(\downarrow\)\\
    \midrule
    TPS*~\cite{zhao2022thin}~~ &\textcolor{lightgray}{3.23E-04}~~ &\textcolor{lightgray}{29.18}~~ & \textcolor{lightgray}{0.673}~~ &\textcolor{lightgray}{0.299}~~ &\textcolor{lightgray}{53.78}~~ & \textcolor{lightgray}{72.55}~~ &\textcolor{lightgray}{306.17}\\
    MRAA*~\cite{siarohin2021motion}~~ &\textcolor{lightgray}{3.21E-04}~~ &\textcolor{lightgray}{29.39}~~ & \textcolor{lightgray}{0.672}~~ &\textcolor{lightgray}{0.296}~~ &\textcolor{lightgray}{54.47}~~ & \textcolor{lightgray}{66.36}~~ &\textcolor{lightgray}{284.82}\\
    \midrule
    TPS~\cite{zhao2022thin}~~ &6.17E-04~~ &28.17~~ &0.560~~ &0.449~~ &140.37~~ &142.52~~ &800.77 \\
    MRAA~\cite{siarohin2021motion}~~ &4.61E-04~~ &28.39~~ &0.646~~ &0.337~~ &85.49~~ &71.97~~ &468.66 \\
    IPA~\cite{ye2023ip}+CtrlN~\cite{zhang2023adding}~~ &7.38E-04~~ &28.03~~ &0.459~~ &0.481~~ &69.83~~ &113.31~~ &802.44 \\
    IPA~\cite{ye2023ip}+CtrlN~\cite{zhang2023adding}-V~~ &6.99E-04~~ &28.00~~ &0.479~~ &0.461~~ &66.81~~ &86.33~~ &666.27 \\
    DisCo~\cite{wang2023disco}~~ &\underline{3.78E-04}~~ &\underline{29.03}~~ &\underline{0.668}~~ &\underline{0.292}~~ &{\bf 30.75}~~ &\underline{59.90}~~ &\underline{292.80} \\
    \ours{} (Ours)~~ &{\bf 3.13E-04}~~ &{\bf 29.16}~~ &{\bf 0.714}~~ &{\bf 0.239}~~ &\underline{32.09}~~ &{\bf 21.75}~~ &{\bf 179.07} \\
    \bottomrule
  \end{tabular}
  \caption{Quantitative comparisons on TikTok~\cite{jafarian2021learning} dataset. We cite results directly from~\cite{wang2023disco} for DisCo, TPS*, and MRAA*.}
  \label{tab:comp:tiktok}
  \end{subtable}
  \vspace{\fill}
  \begin{subtable}[t]{\linewidth}
  \centering
  \begin{tabular}{lccccccc}
    \toprule
    \multirow{2}{*}{Method} &\multicolumn{5}{c}{Image} &\multicolumn{2}{c}{Video}\\
    \cmidrule(r){2-6} \cmidrule(r){7-8} &AKD\(\downarrow\) &MKR\(\downarrow\)  &AED\(\downarrow\)& (L1, L1\(_{\text{fg}}\))\(\downarrow\)  & FID\(\downarrow\) & FID-VID\(\downarrow\)& FVD\(\downarrow\)\\
    \midrule
    TPS*~\cite{zhao2022thin} &\textcolor{lightgray}{2.16} &\textcolor{lightgray}{0.008} & \textcolor{lightgray}{0.167} &\textcolor{lightgray}{(1.04E-04, 7.24E-05)} &\textcolor{lightgray}{22.35} &\textcolor{lightgray}{6.88} & \textcolor{lightgray}{81.41}\\
    MRAA*~\cite{siarohin2021motion} &\textcolor{lightgray}{2.50}&\textcolor{lightgray}{0.007} & \textcolor{lightgray}{0.154} &\textcolor{lightgray}{(1.06E-04, 7.12E-05)} &\textcolor{lightgray}{21.13} &\textcolor{lightgray}{6.03} & \textcolor{lightgray}{78.29}\\
    \midrule
    TPS~\cite{zhao2022thin} &11.00 &0.063 &0.331 &(2.22E-04, 1.43E-04) &86.65 &72.49 &457.02 \\
    MRAA~\cite{siarohin2021motion} &4.37 &0.024 &\underline{0.246} &({\bf 1.61E-04}, {\bf 1.07E-04}) &35.75 &22.97 &\underline{182.78} \\
    IPA~\cite{ye2023ip}+CtrlN~\cite{zhang2023adding} &5.14 &0.022 &0.375 &(3.58E-04, 1.71E-04) &43.23 &49.13 &434.00 \\
    IPA~\cite{ye2023ip}+CtrlN~\cite{zhang2023adding}-V &4.24 &\underline{0.019} &0.369 &(4.43E-04, 1.66E-04) &49.21 &38.48 &281.42 \\
    DisCo~\cite{wang2023disco} &\underline{2.96} &\underline{0.019} &0.253 &(\underline{2.07E-04}, 1.31E-04) &\underline{27.51} &\underline{19.02} &195.00 \\
    \ours{} (Ours) &{\bf 2.65} &{\bf 0.013} &{\bf 0.204} &(2.92E-04, \underline{1.11E-04}) &{\bf 22.78} &{\bf 19.00} &{\bf 131.51} \\
    \bottomrule
  \end{tabular}
  \caption{Quantitative comparisons on TED-talks~\cite{siarohin2021motion} dataset. We also report L1 error for the foreground (human regions).}
  \label{tab:comp:tedtalks}
  \end{subtable}
  \vspace{-0.9em}
  \caption{Quantitative comparisons with baselines, with best results in {\bf bold} and second best results \underline{underlined}. *The original TPS and MRAA directly use ground-truth video frames for animation, we report their results (marked in gray) only for reference.}
  \label{tab:comp}
\vspace{-1.2em}
\end{table*}
\subsection{Comparisons}
\noindent {\bf Baselines.} We perform a comprehensive comparison of \ours{} with several state-of-the-art methods for human image animation: (1) {\bf MRAA}~\cite{siarohin2021motion} and {\bf TPS}~\cite{zhao2022thin} are state-of-the-art GAN-based animation approaches, which estimate optical flow from driving sequences to warp the source image and then inpaint the occluded regions using GAN models. (2) {\bf DisCo}~\cite{wang2023disco} is the state-of-the-art diffusion-based animation method that integrates disentangled condition modules for pose, human, and background into a pretrained diffusion model to implement human image animation.
(3) We construct additional baseline by combining the state-of-the-art image condition method, \ie, IP-Adapter~\cite{ye2023ip}, with pose ControlNet~\cite{zhang2023adding}, which is labeled as {\bf IPA+CtrlN}. To make a fair comparison, we further add temporal attention blocks~\cite{guo2023animatediff} into this framework and construct a video version baseline labeled as {\bf IPA+CtrlN-V}.
In addition, MRAA and TPS methods utilize ground-truth videos as driving signals. To ensure fair comparisons, we train alternative versions for MRAA and TPS using the same driving signal (DensePose) as \ours{}.

\noindent {\bf Evaluation metrics.} We adhere to established evaluation metrics employed in prior research. For the TikTok dataset, we evaluate both single-frame image quality and video fidelity. The metrics used for single-frame quality include L1 error, SSIM~\cite{wang2004image}, LPIPS~\cite{zhang2018unreasonable}, PSNR~\cite{hore2010image}, and FID~\cite{heusel2017gans}. Video fidelity is assessed through FID-FVD~\cite{balaji2019conditional} and FVD~\cite{unterthiner2018towards}. On the TED-talks dataset, we follow MRAA and report L1 error, average keypoint distance (AKD), missing keypoint rate (MKR), and average Euclidean distance (AED). However, these evaluation metrics are designed for single-frame evaluation and lack perceptual measurement of the animation results. Consequently, we also compute FID, FID-VID, and FVD on the TED-talks dataset to measure the image and video perceptual quality.

\begin{figure*}[t]
\centering
\includegraphics[width=0.86\textwidth]{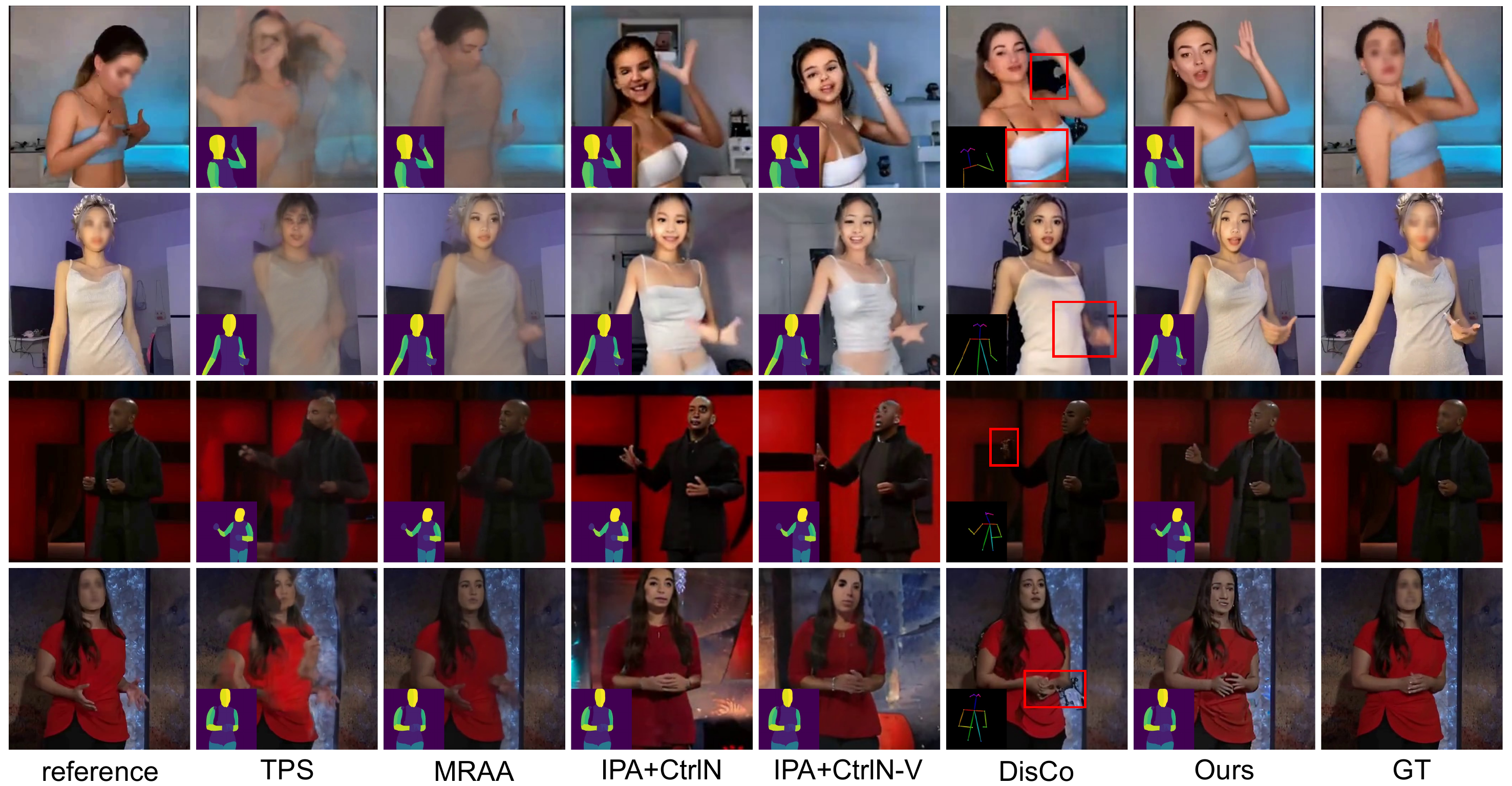}
\vspace{-0.9em}
\caption{
Qualitative comparisons between \ours{} and baselines on TikTok and TED-talks datasets. We overlay the target pose on the bottom left corner of the synthesized frames and highlight the artifacts generated by the strongest baseline (DisCo) in red boxes. For comprehensive video comparisons, please refer to our \href{\projpage}{Project Page}.
}
\label{fig:qual}
\vspace{-1.4em}
\end{figure*}
\noindent {\bf Quantitative comparisons.}
Table~\ref{tab:comp} provides the quantitative comparison results between \ours{} and baselines on two benchmark datasets. Our method surpasses all baselines in terms of reconstruction metrics, \ie, L1, PSNR, SSIM, and LPIPS, on TikTok (Table~\ref{tab:comp:tiktok}). Notably, \ours{} improves against the strongest baseline (DisCo) by 6.9\% and 18.2\% for SSIM and LPIPS, respectively. 
Additionally, \ours{} achieves state-of-the-art video fidelity, demonstrating significant performance improvements of 63.7\% for FID-VID and 38.8\% for FVD compared to DisCo.

Our method also exhibits superior video fidelity on the TED-talks dataset (Table~\ref{tab:comp:tedtalks}), achieving the best FID-VID of 19.00 and FVD of 131.51. This performance is particularly notable against the second-best method (MRAA), with an improvement of 28.1\% for FVD. Additionally, \ours{} demonstrates state-of-the-art single-frame fidelity, securing the best FID score of 22.78. Compared with DisCo, a diffusion-based baseline method, \ours{} showcases a significant improvement of 17.2\%.
However, it is important to note that \ours{} has a higher L1 error compared to baselines. This is likely caused by the lack of background information in the DensePose control signals. Consequently, \ours{} is unable to learn a consistent dynamic background as presented in the TED-talks dataset, leading to an increased L1 error. Nevertheless, \ours{} achieves a comparable L1 error with the strongest baseline (MRAA) in foreground human regions, demonstrating its effectiveness for human animations.
Furthermore, our method achieves the best performance for AKD, MKR, and AED, providing evidence of its superior identity-preserving ability and animation precision.

\noindent {\bf Qualitative comparisons.}
In Figure~\ref{fig:qual}, we present qualitative comparisons between \ours{} and baselines. Notably, the dancing videos from the TikTok dataset exhibit significant pose variations, posing a challenge for GAN-based methods such as MRAA and TPS, as they struggle to produce reasonable results when there is a substantial pose difference between the reference image and the driving signal. In contrast, the diffusion-based baselines, IPA+CtrlN, IPA+CtrlN-V, and DisCo, show better single-frame quality. 
However, as IPA+CtrlN and DisCo generate each frame independently, their temporal consistency is unsatisfactory, as evidenced by the color change of the clothes and inconsistent backgrounds in the occluded regions. The video diffusion baseline, IPA+CtrlN-V, displays more consistent content, yet its single-frame quality is inferior due to weak reference conditioning. Conversely, \ours{} produces temporally consistent animations and high-fidelity details for the background, clothes, face, and hands.

Unlike the TikTok dataset, TED-talks dataset comprises speech videos recorded under dim lighting conditions. The motions in the TED-talks dataset primarily involve gestures, which are less challenging than dancing videos. Thus, MRAA and TPS  produce more visually plausible results, albeit with inaccurate motion. In contrast, IPA+CtrlN, IPA+CtrlN-V, DisCo, and  \ours{} demonstrate a more precise body pose control ability because these methods extract appearance conditions from the reference image to guide the animation instead of directly warping the source image. Among all these methods, \ours{} exhibits superior identity- and background-preserving ability, as shown in Figure~\ref{fig:qual}, thanks to our appearance encoder, which extracts detailed information from reference image.

\noindent{\bf Cross-identity animation.}
Beyond animating each identity with its corresponding motion sequence, we further investigate the cross-identity animation capability of \ours{} and the state-of-the-art baselines, \ie, DisCo, and MRAA. 
Specifically, we sample two DensePose motion sequences from the TikTok test set and use these sequences to animate reference images from other videos.
Figure~\ref{fig:app1:crossid} illustrates that MRAA fails to generalize for driving videos that contain substantial pose differences, while DisCo struggles to preserve the details in the reference images, resulting in artifacts in the background and clothing. In contrast, our method faithfully animates the reference images given the target motion, demonstrating its robustness.

\subsection{Ablation Studies}
To verify the effectiveness of the design choices in \ours{}, we conduct ablative experiments on the TikTok dataset, which features significant pose variations, a wide range of identities, and diverse backgrounds.

\begin{figure}[t]
\centering
\begin{subfigure}[b]{0.87\linewidth}
 \centering
 \includegraphics[width=\textwidth]{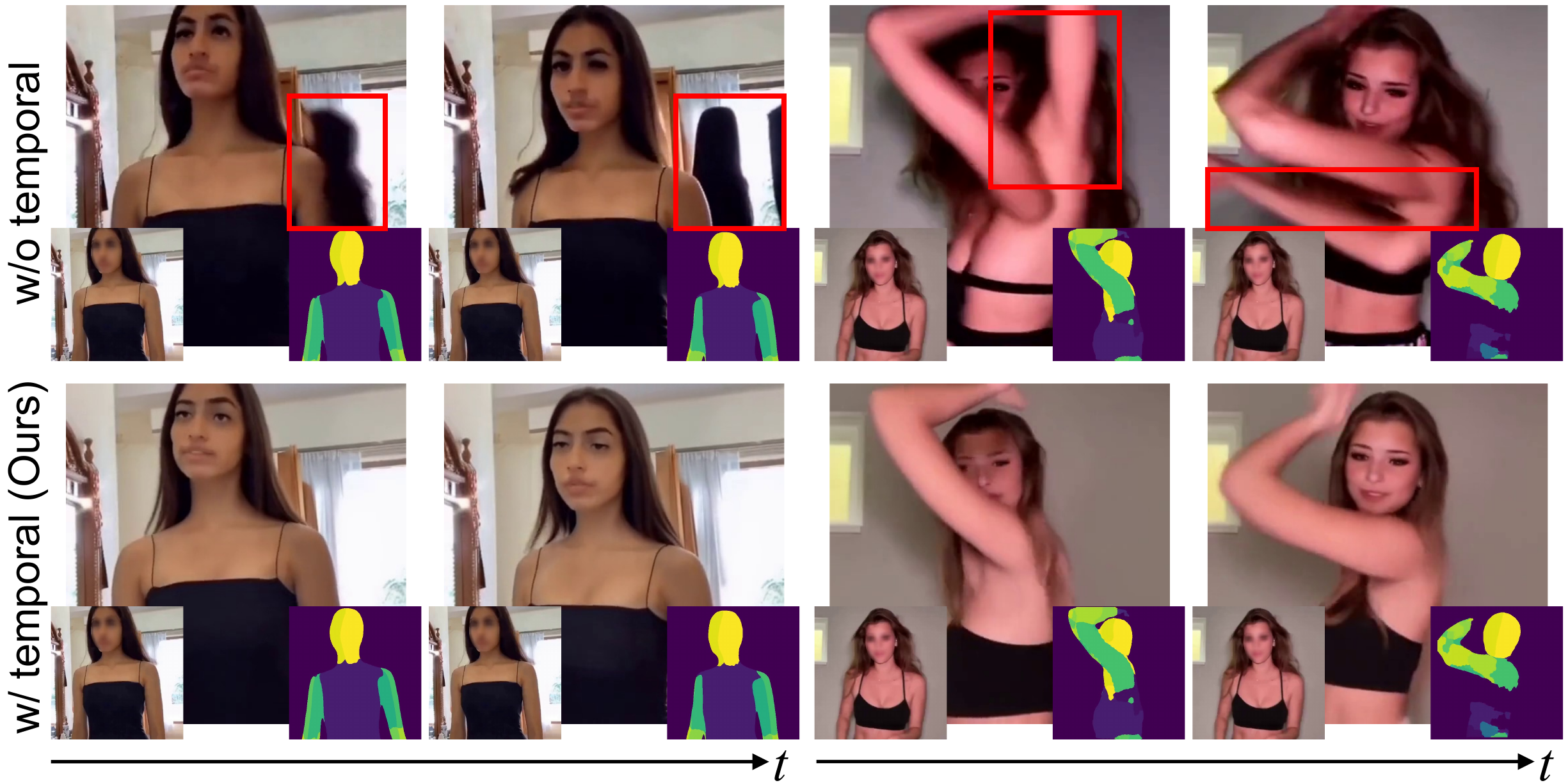}
 \caption{Effects of temporal modeling.}
 \label{fig:ab:temp}
\end{subfigure}
\hfill
\begin{subfigure}[b]{0.87\linewidth}
 \centering
 \includegraphics[width=\textwidth]{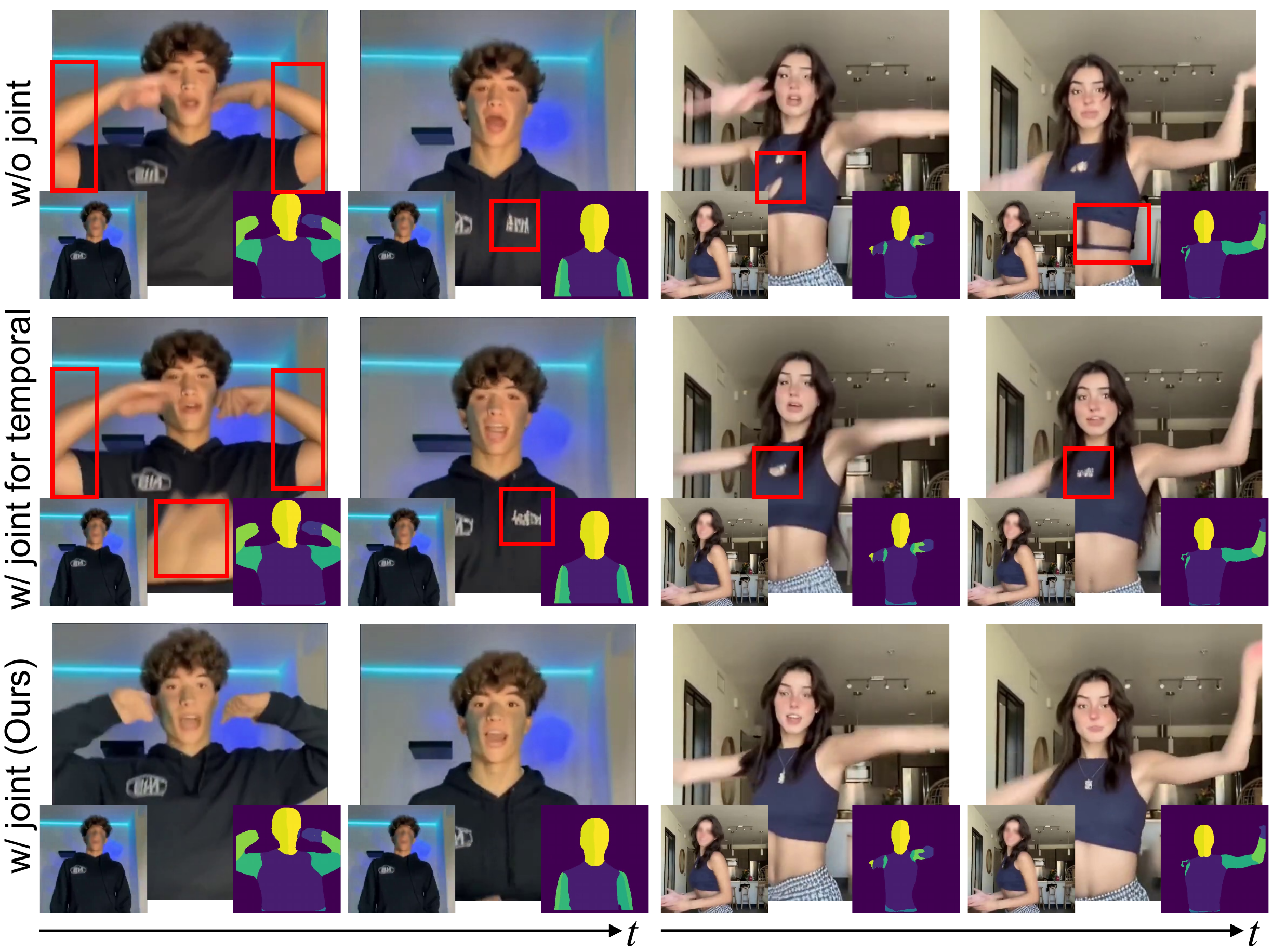}
 \caption{Effects of image-video joint training strategy.}
 \label{fig:ab:joint}
\end{subfigure}
\vspace{-0.5em}
\caption{Visualization of ablation studies, with errors highlighted in red boxes. For each frame, we overlay the reference image at the bottom left corner, and the target pose at the bottom right corner.}
\label{fig:ab}
\vspace{-1.8em}
\end{figure}

\noindent{\bf Temporal modeling.}
To assess the impact of the proposed temporal attention layer, we train a version of \ours{} {\it without} it for comparison. The results, presented in Table~\ref{tab:ab:temp_attn}, show a decrease in both single-frame quality and video fidelity evaluation metrics when the temporal attention layers are discarded, highlighting the effectiveness of our temporal modeling.
This is further supported by the qualitative ablation results presented in Figure~\ref{fig:ab:temp}, where the model without explicitly temporal modeling fails to maintain temporal coherence for both humans and backgrounds.

\begin{table}[t]
  \renewcommand{\tabcolsep}{0.6pt}
  \small
\begin{subtable}[t]{\linewidth}
    \centering
  \begin{tabular}{cccccccc}
    \toprule
    {\it Temp Attn}~&L1\(\downarrow\)  &PSNR\(\uparrow\)    &SSIM\(\uparrow\)  &LPIPS\(\downarrow\) &FID\(\downarrow\) &FID-VID\(\downarrow\) &FVD\(\downarrow\)\\
    \midrule
    w/o &3.98 &28.90 &0.652 &0.263 &{\bf 27.54} &42.21 &247.30\\
    w/ &{\bf 3.13} &{\bf 29.16} &{\bf 0.714} &{\bf 0.239} &32.09 &{\bf 21.75} &{\bf 179.07}\\
    \bottomrule
  \end{tabular}
  \caption{The effect of modeling temporal information.}
  \label{tab:ab:temp_attn}
\end{subtable}
\hspace{\fill}
\begin{subtable}[t]{\linewidth}
    \centering
  \begin{tabular}{ccccccccc}
    \toprule
    {\it App Enc}&L1\(\downarrow\)  &PSNR\(\uparrow\)    &SSIM\(\uparrow\)  &LPIPS\(\downarrow\) &FID\(\downarrow\) &FID-VID\(\downarrow\) &FVD\(\downarrow\)\\
    \midrule
    CLIP &8.00 &27.94 &0.461 &0.481 &78.35 &82.50 &724.96\\
    IP-Adapter &7.89 &27.98 &0.481 &0.442 &64.17 &67.65 &590.99\\
    Ours &{\bf 3.13} &{\bf 29.16} &{\bf 0.714} &{\bf 0.239} &{\bf 32.09} &{\bf 21.75} &{\bf 179.07}\\
    \bottomrule
  \end{tabular}
  \caption{The effect of appearance encoder.}
  \label{tab:ab:ap_enc}
\end{subtable}
\hspace{\fill}
\begin{subtable}[t]{\linewidth}
    \centering
  \begin{tabular}{ccccccccc}
    \toprule
    {\it Spat}~~&{\it Temp}~ &L1\(\downarrow\)  &PSNR\(\uparrow\)    &SSIM\(\uparrow\)  &LPIPS\(\downarrow\) &FID\(\downarrow\) &FID-VID\(\downarrow\) &FVD\(\downarrow\)\\
    \midrule
    \xmark &\xmark &3.20 &29.09 &0.706 &0.248 &37.15 &24.45 &158.16 \\
    \xmark &\cmark &3.19 &29.12 &0.705 &0.246 &38.41 &23.08 &{\bf 156.32} \\
    \cmark &\cmark &{\bf 3.13} &{\bf 29.16} &{\bf 0.714} &{\bf 0.239} &{\bf 32.09} &{\bf 21.75} &179.07 \\
    \bottomrule
  \end{tabular}
  \caption{The effect of image-video joint training.}
  \label{tab:ab:joint}
\end{subtable}
\hspace{\fill}
\begin{subtable}[t]{0.49\linewidth}
    \centering
  \begin{tabular}{cccc}
    \toprule
    {\it Avg}~ &L1\(\downarrow\)~ &FID\(\downarrow\) &FID-FVD\(\downarrow\)\\
    \midrule
    w/o~ &3.21~ &32.99 &22.50\\
    w/~ &{\bf 3.13}~ &{\bf 32.08} &{\bf 21.75}\\
    \bottomrule
  \end{tabular}
  \caption{The effect of the inference-stage temporal video fusion.}
  \label{tab:ab:overlap}
\end{subtable}
\vspace{\fill}
\begin{subtable}[t]{0.5\linewidth}
    \centering
  \begin{tabular}{cccc}
    \toprule
    {\it Noise}~ &L1\(\downarrow\)~ &FID\(\downarrow\) &FID-FVD\(\downarrow\) \\
    \midrule
    diff~ &{\bf 3.03}~ &32.74 &22.50 \\
    same~ &3.13~ &{\bf 32.08} &{\bf 21.75} \\
    \bottomrule
  \end{tabular}
  \caption{The effect of sharing the same initial noises for all the video segments.}
  \label{tab:ab:noise}
\end{subtable}
\vspace{-0.5em}
\caption{Ablations of \ours{} on TikTok dataset, with best results in {\bf bold}. We vary our architectural designs and training strategies to investigate their effectiveness. We report L1\(\times10^{-4}\) for numerical simplicity.}
\label{tab:ab}
\vspace{-1em}
\end{table}

\noindent{\bf Appearance encoder.} To evaluate the enhancement brought by the proposed appearance encoding strategy, we replace the appearance encoder in \ours{} with CLIP~\cite{radford2021learning} and IP-Adapter~\cite{ye2023ip} to establish baselines.
Table~\ref{tab:ab:ap_enc} summarizes the ablative results. It is evident that our method significantly outperforms these two baselines in reference image preserving, resulting in a substantial improvement for both single-frame and video fidelity.

\noindent{\bf Inference-stage video fusion.} \ours{} utilizes a video fusion technique to enhance the transition smoothness of long-term animation. Table~\ref{tab:ab:overlap} and Table~\ref{tab:ab:noise} demonstrate the effectiveness of our design choices. In general, skipping the video fusion or using different initial random noises for different video segments diminishes animation performance, as evidenced by the performance drop for both appearance and video quality.

\noindent{\bf Image-video joint training.} 
We introduce an image-video joint training strategy to enhance the animation quality. As shown in Table~\ref{tab:ab:joint},  applying image-video joint training at both the appearance encoding and temporal modeling stages consistently increases the animation quality. Such improvements can also be observed in Figure~\ref{fig:ab:joint}.
Without the joint training strategy, the model struggles to model intricate details accurately, tending to produce incorrect clothes and accessories as shown in Figure~\ref{fig:ab:joint}.

\begin{figure}[t]
\centering
\begin{subfigure}[b]{0.88\linewidth}
 \centering
 \includegraphics[width=\textwidth]{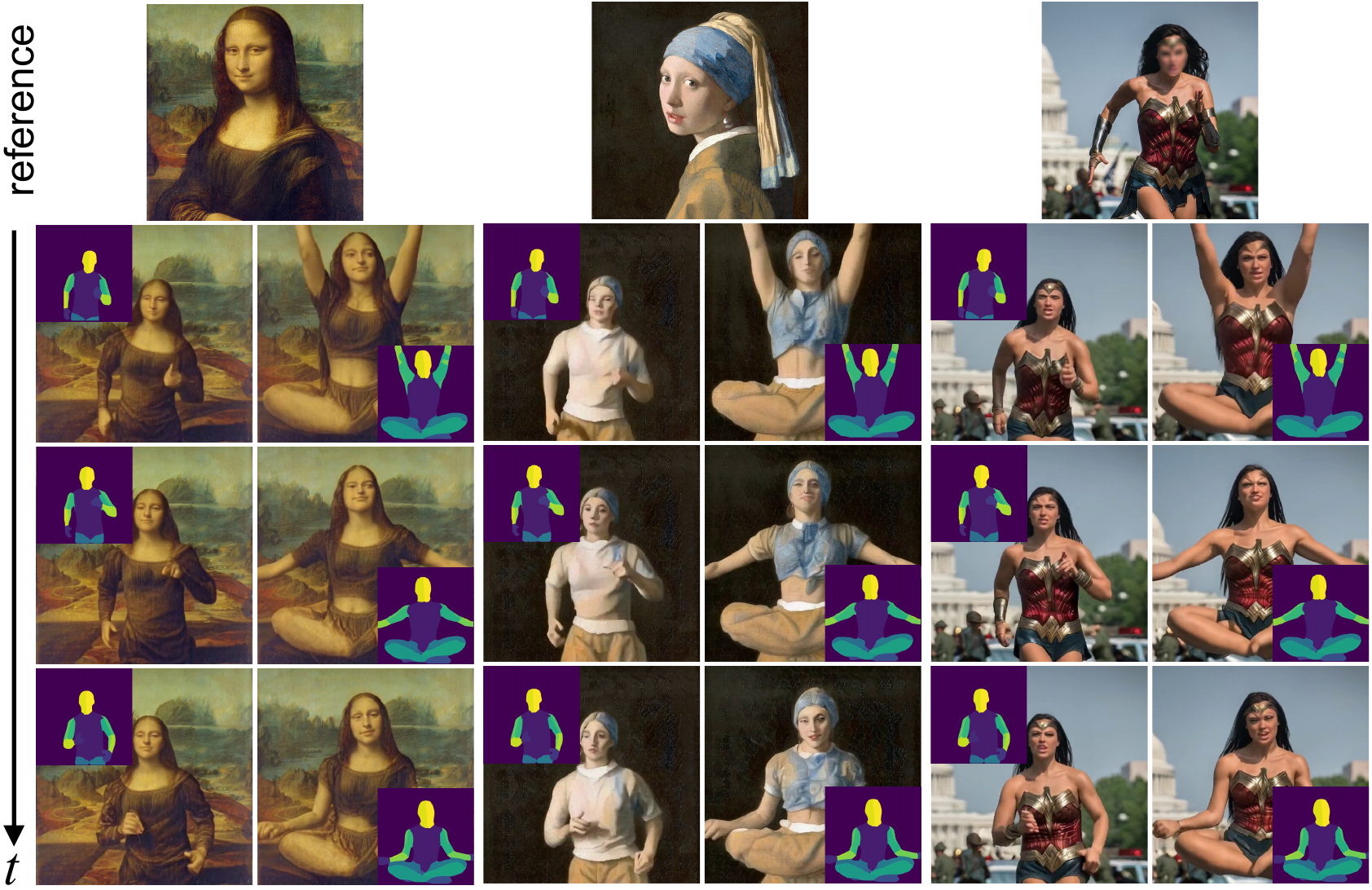}
 \caption{Unseen domain animation.}
 \label{fig:app1:unseen}
 \vspace{0.3em}
\end{subfigure}
\vfill
\begin{subfigure}[b]{0.88\linewidth}
 \centering
 \includegraphics[width=\textwidth]{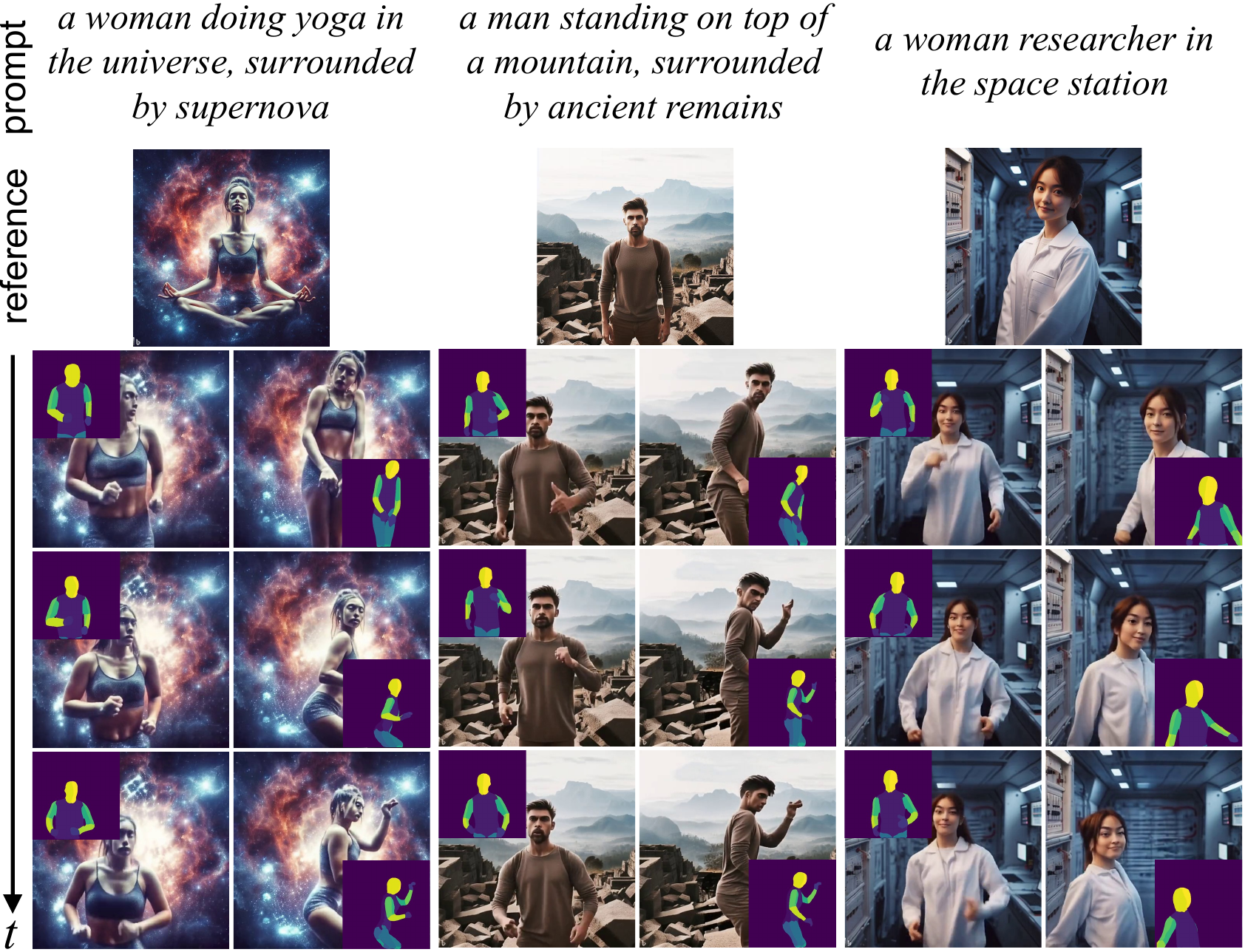}
 \caption{Combining \ours{} with T2I diffusion model.}
 \label{fig:app2:dalle}
\end{subfigure}
\vfill
\begin{subfigure}[b]{0.88\linewidth}
 \centering
 \includegraphics[width=\textwidth]{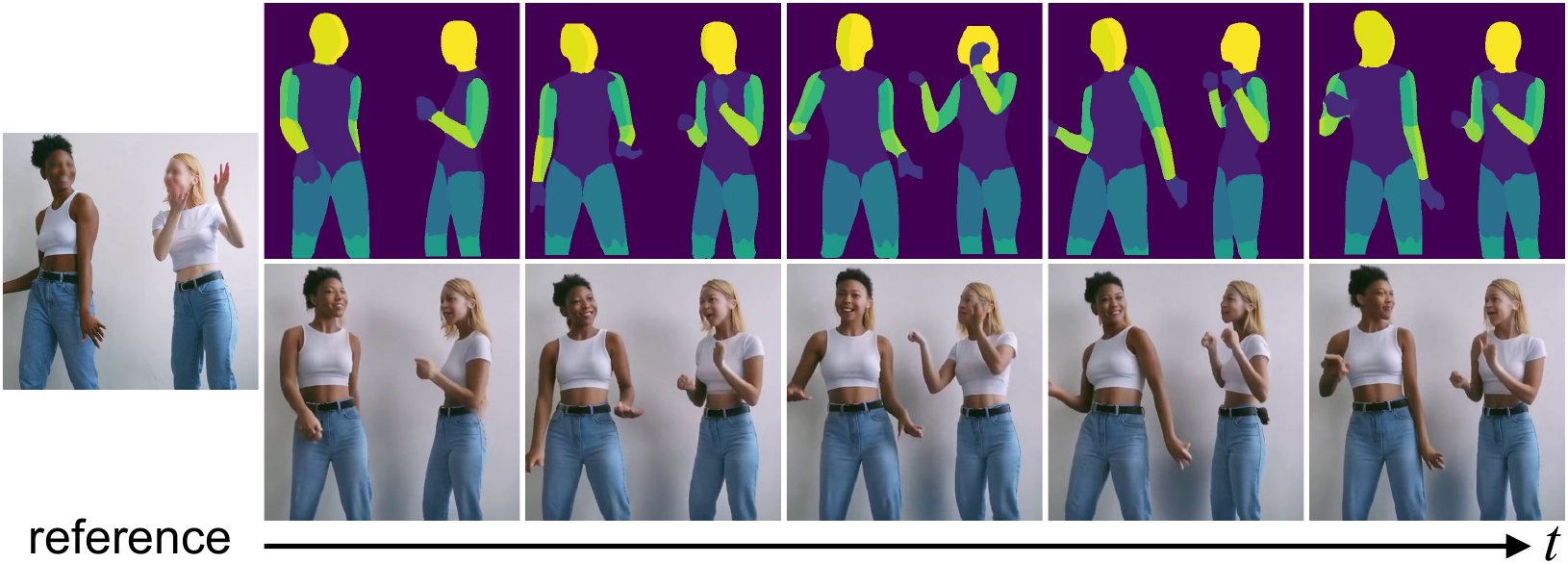}
 \caption{Multi-person animation.}
 \label{fig:app2:multi}
\end{subfigure}
\vspace{-0.9em}
\caption{(a) Animation results for the unseen domain. (b) Combining \ours{} with DALL\(\cdot\)E3~\cite{dalle3}, and (c) Multi-person animation. We overlay the motion signal at the corner of each frame in (a) and (b). Video results can be found on \href{\projpage}{Project Page}.}
\label{fig:app2}
\vspace{-1em}
\end{figure}
\subsection{Applications}
Despite being trained only on realistic human data, \ours{} demonstrates the ability to generalize to various application scenarios, including animating unseen domain data, integration with a text-to-image diffusion model, and multi-person animation.

\noindent{\bf Unseen domain animation.}
\ours{} showcases generalization ability for unseen image styles and motion sequences. As shown in Figure~\ref{fig:app1:unseen}, it can animate oil paintings and movie images to perform actions such as running and Yoga, 
{maintaining a stable background and inpainting the occluded regions with temporally consistent results.}

\noindent{\bf Combining with text-to-image generation.}
Due to its strong generalization ability, \ours{} can be used to animate images generated by text-to-image (T2I) models, \eg, DALL\(\cdot\)E3~\cite{dalle3}. As shown in Figure~\ref{fig:app2:dalle}, we first employ DALL\(\cdot\)E3 to synthesize the reference image using various prompts. These reference images can then be animated by our model to perform various actions.

\noindent{\bf Multi-person animation.}
\ours{} also exhibits strong generalization for multi-person animation. As illustrated in Figure~\ref{fig:app2:multi}, we can generate animations for multiple individuals given the reference frame and a motion sequence, which includes two dancing individuals.

%% file: sec/6_conclusion.tex
\section{Conclusion}
\label{conclusion}
This work introduces \ours{}, a novel diffusion-based framework designed for human avatar animation with an emphasis on temporal consistency. By effectively modeling temporal information, we enhance the overall temporal coherence of the animation results. The proposed appearance encoder not only elevates single-frame quality but also contributes to improved temporal consistency. Additionally, the integration of a video frame fusion technique enables seamless transitions across the animation video. \ours{} demonstrates state-of-the-art performance in terms of both single-frame and video quality. Moreover, its robust generalization capabilities make it applicable to unseen domains and multi-person animation scenarios.

%% file: sec/7_acknowledgement.tex
\section*{Acknowledgement}
This project is supported by the National Research Foundation, Singapore under its NRFF Award NRF-NRFF13-2021-0008, and the Ministry of Education, Singapore, under the Academic Research Fund Tier 1 (FY2022).

%% file: main.bbl
\begin{thebibliography}{52}
\providecommand{\natexlab}[1]{#1}
\providecommand{\url}[1]{\texttt{#1}}
\expandafter\ifx\csname urlstyle\endcsname\relax
  \providecommand{\doi}[1]{doi: #1}\else
  \providecommand{\doi}{doi: \begingroup \urlstyle{rm}\Url}\fi

\bibitem[Balaji et~al.(2019)Balaji, Min, Bai, Chellappa, and Graf]{balaji2019conditional}
Yogesh Balaji, Martin~Renqiang Min, Bing Bai, Rama Chellappa, and Hans~Peter Graf.
\newblock Conditional gan with discriminative filter generation for text-to-video synthesis.
\newblock In \emph{IJCAI}, 2019.

\bibitem[Betker et~al.(2023)Betker, Goh, Jing, Brooks, Wang, Li, Ouyang, Zhuang, Lee, Guo, Manassra, Dhariwal, Chu, and Jiao]{dalle3}
James Betker, Gabriel Goh, Li Jing, Tim Brooks, Jianfeng Wang, Linjie Li, Long Ouyang, Juntang Zhuang, Joyce Lee, Yufei Guo, Wesam Manassra, Prafulla Dhariwal, Casey Chu, and Yunxin Jiao.
\newblock Improving image generation with better captions.
\newblock \url{https://cdn.openai.com/papers/dall-e-3.pdf}, 2023.

\bibitem[Cao et~al.(2014)Cao, Hou, and Zhou]{cao2014displaced}
Chen Cao, Qiming Hou, and Kun Zhou.
\newblock Displaced dynamic expression regression for real-time facial tracking and animation.
\newblock \emph{ACM TOG}, 2014.

\bibitem[Cao et~al.(2023)Cao, Wang, Qi, Shan, Qie, and Zheng]{cao2023masactrl}
Mingdeng Cao, Xintao Wang, Zhongang Qi, Ying Shan, Xiaohu Qie, and Yinqiang Zheng.
\newblock Masactrl: Tuning-free mutual self-attention control for consistent image synthesis and editing.
\newblock In \emph{ICCV}, 2023.

\bibitem[Cao et~al.(2017)Cao, Simon, Wei, and Sheikh]{cao2017realtime}
Zhe Cao, Tomas Simon, Shih-En Wei, and Yaser Sheikh.
\newblock Realtime multi-person 2d pose estimation using part affinity fields.
\newblock In \emph{CVPR}, 2017.

\bibitem[Chan et~al.(2019)Chan, Ginosar, Zhou, and Efros]{chan2019everybody}
Caroline Chan, Shiry Ginosar, Tinghui Zhou, and Alexei~A Efros.
\newblock Everybody dance now.
\newblock In \emph{CVPR}, 2019.

\bibitem[Geng et~al.(2019)Geng, Cao, and Tulyakov]{geng20193d}
Zhenglin Geng, Chen Cao, and Sergey Tulyakov.
\newblock 3d guided fine-grained face manipulation.
\newblock In \emph{CVPR}, 2019.

\bibitem[G{\"u}ler et~al.(2018)G{\"u}ler, Neverova, and Kokkinos]{guler2018densepose}
R{\i}za~Alp G{\"u}ler, Natalia Neverova, and Iasonas Kokkinos.
\newblock Densepose: Dense human pose estimation in the wild.
\newblock In \emph{CVPR}, 2018.

\bibitem[Guo et~al.(2019)Guo, Lincoln, Davidson, Busch, Yu, Whalen, Harvey, Orts-Escolano, Pandey, Dourgarian, et~al.]{guo2019relightables}
Kaiwen Guo, Peter Lincoln, Philip Davidson, Jay Busch, Xueming Yu, Matt Whalen, Geoff Harvey, Sergio Orts-Escolano, Rohit Pandey, Jason Dourgarian, et~al.
\newblock The relightables: Volumetric performance capture of humans with realistic relighting.
\newblock \emph{ACM TOG}, 2019.

\bibitem[Guo et~al.(2023)Guo, Yang, Rao, Wang, Qiao, Lin, and Dai]{guo2023animatediff}
Yuwei Guo, Ceyuan Yang, Anyi Rao, Yaohui Wang, Yu Qiao, Dahua Lin, and Bo Dai.
\newblock Animatediff: Animate your personalized text-to-image diffusion models without specific tuning.
\newblock \emph{arXiv}, 2023.

\bibitem[Heusel et~al.(2017)Heusel, Ramsauer, Unterthiner, Nessler, and Hochreiter]{heusel2017gans}
Martin Heusel, Hubert Ramsauer, Thomas Unterthiner, Bernhard Nessler, and Sepp Hochreiter.
\newblock Gans trained by a two time-scale update rule converge to a local nash equilibrium.
\newblock In \emph{NeurIPS}, 2017.

\bibitem[Hong et~al.(2023)Hong, Chen, LAN, Pan, and Liu]{hong2023evad}
Fangzhou Hong, Zhaoxi Chen, Yushi LAN, Liang Pan, and Ziwei Liu.
\newblock {EVA}3d: Compositional 3d human generation from 2d image collections.
\newblock In \emph{ICLR}, 2023.

\bibitem[Hore and Ziou(2010)]{hore2010image}
Alain Hore and Djemel Ziou.
\newblock Image quality metrics: Psnr vs. ssim.
\newblock In \emph{ICPR}, 2010.

\bibitem[Jafarian and Park(2021)]{jafarian2021learning}
Yasamin Jafarian and Hyun~Soo Park.
\newblock Learning high fidelity depths of dressed humans by watching social media dance videos.
\newblock In \emph{CVPR}, 2021.

\bibitem[Karras et~al.(2023)Karras, Holynski, Wang, and Kemelmacher-Shlizerman]{karras2023dreampose}
Johanna Karras, Aleksander Holynski, Ting-Chun Wang, and Ira Kemelmacher-Shlizerman.
\newblock Dreampose: Fashion image-to-video synthesis via stable diffusion.
\newblock \emph{arXiv}, 2023.

\bibitem[Ma et~al.(2023)Ma, He, Cun, Wang, Shan, Li, and Chen]{ma2023follow}
Yue Ma, Yingqing He, Xiaodong Cun, Xintao Wang, Ying Shan, Xiu Li, and Qifeng Chen.
\newblock Follow your pose: Pose-guided text-to-video generation using pose-free videos.
\newblock \emph{arXiv}, 2023.

\bibitem[Mallya et~al.(2022)Mallya, Wang, and Liu]{mallya2022implicit}
Arun Mallya, Ting-Chun Wang, and Ming-Yu Liu.
\newblock Implicit warping for animation with image sets.
\newblock In \emph{NeurIPS}, 2022.

\bibitem[Nirkin et~al.(2019)Nirkin, Keller, and Hassner]{nirkin2019fsgan}
Yuval Nirkin, Yosi Keller, and Tal Hassner.
\newblock Fsgan: Subject agnostic face swapping and reenactment.
\newblock In \emph{ICCV}, 2019.

\bibitem[Oorloff and Yacoob(2023)]{oorloff2023robust}
Trevine Oorloff and Yaser Yacoob.
\newblock Robust one-shot face video re-enactment using hybrid latent spaces of stylegan2.
\newblock In \emph{ICCV}, 2023.

\bibitem[Qian et~al.(2019)Qian, Lin, Wu, Liu, Wang, Shen, Qian, and He]{qian2019make}
Shengju Qian, Kwan-Yee Lin, Wayne Wu, Yangxiaokang Liu, Quan Wang, Fumin Shen, Chen Qian, and Ran He.
\newblock Make a face: Towards arbitrary high fidelity face manipulation.
\newblock In \emph{ICCV}, 2019.

\bibitem[Radford et~al.(2021)Radford, Kim, Hallacy, Ramesh, Goh, Agarwal, Sastry, Askell, Mishkin, Clark, et~al.]{radford2021learning}
Alec Radford, Jong~Wook Kim, Chris Hallacy, Aditya Ramesh, Gabriel Goh, Sandhini Agarwal, Girish Sastry, Amanda Askell, Pamela Mishkin, Jack Clark, et~al.
\newblock Learning transferable visual models from natural language supervision.
\newblock In \emph{ICML}, 2021.

\bibitem[Ren et~al.(2021)Ren, Li, Chen, Li, and Liu]{ren2021pirenderer}
Yurui Ren, Ge Li, Yuanqi Chen, Thomas~H Li, and Shan Liu.
\newblock Pirenderer: Controllable portrait image generation via semantic neural rendering.
\newblock In \emph{ICCV}, 2021.

\bibitem[Rombach et~al.(2022)Rombach, Blattmann, Lorenz, Esser, and Ommer]{rombach2022high}
Robin Rombach, Andreas Blattmann, Dominik Lorenz, Patrick Esser, and Bj{\"o}rn Ommer.
\newblock High-resolution image synthesis with latent diffusion models.
\newblock In \emph{CVPR}, 2022.

\bibitem[Saharia et~al.(2022)Saharia, Chan, Saxena, Li, Whang, Denton, Ghasemipour, Gontijo~Lopes, Karagol~Ayan, Salimans, et~al.]{saharia2022photorealistic}
Chitwan Saharia, William Chan, Saurabh Saxena, Lala Li, Jay Whang, Emily~L Denton, Kamyar Ghasemipour, Raphael Gontijo~Lopes, Burcu Karagol~Ayan, Tim Salimans, et~al.
\newblock Photorealistic text-to-image diffusion models with deep language understanding.
\newblock In \emph{NeurIPS}, 2022.

\bibitem[Schuhmann et~al.(2021)Schuhmann, Vencu, Beaumont, Kaczmarczyk, Mullis, Katta, Coombes, Jitsev, and Komatsuzaki]{schuhmann2021laion}
Christoph Schuhmann, Richard Vencu, Romain Beaumont, Robert Kaczmarczyk, Clayton Mullis, Aarush Katta, Theo Coombes, Jenia Jitsev, and Aran Komatsuzaki.
\newblock Laion-400m: Open dataset of clip-filtered 400 million image-text pairs.
\newblock \emph{arXiv}, 2021.

\bibitem[Siarohin et~al.(2019{\natexlab{a}})Siarohin, Lathuili{\`e}re, Tulyakov, Ricci, and Sebe]{siarohin2019animating}
Aliaksandr Siarohin, St{\'e}phane Lathuili{\`e}re, Sergey Tulyakov, Elisa Ricci, and Nicu Sebe.
\newblock Animating arbitrary objects via deep motion transfer.
\newblock In \emph{CVPR}, 2019{\natexlab{a}}.

\bibitem[Siarohin et~al.(2019{\natexlab{b}})Siarohin, Lathuili{\`e}re, Tulyakov, Ricci, and Sebe]{siarohin2019first}
Aliaksandr Siarohin, St{\'e}phane Lathuili{\`e}re, Sergey Tulyakov, Elisa Ricci, and Nicu Sebe.
\newblock First order motion model for image animation.
\newblock 2019{\natexlab{b}}.

\bibitem[Siarohin et~al.(2021)Siarohin, Woodford, Ren, Chai, and Tulyakov]{siarohin2021motion}
Aliaksandr Siarohin, Oliver Woodford, Jian Ren, Menglei Chai, and Sergey Tulyakov.
\newblock Motion representations for articulated animation.
\newblock In \emph{CVPR}, 2021.

\bibitem[Siarohin et~al.(2023)Siarohin, Menapace, Skorokhodov, Olszewski, Ren, Lee, Chai, and Tulyakov]{siarohin2023unsupervised}
Aliaksandr Siarohin, Willi Menapace, Ivan Skorokhodov, Kyle Olszewski, Jian Ren, Hsin-Ying Lee, Menglei Chai, and Sergey Tulyakov.
\newblock Unsupervised volumetric animation.
\newblock In \emph{CVPR}, 2023.

\bibitem[Song et~al.(2021)Song, Meng, and Ermon]{song2020denoising}
Jiaming Song, Chenlin Meng, and Stefano Ermon.
\newblock Denoising diffusion implicit models.
\newblock 2021.

\bibitem[Thies et~al.(2016)Thies, Zollhofer, Stamminger, Theobalt, and Nie{\ss}ner]{thies2016face2face}
Justus Thies, Michael Zollhofer, Marc Stamminger, Christian Theobalt, and Matthias Nie{\ss}ner.
\newblock Face2face: Real-time face capture and reenactment of rgb videos.
\newblock In \emph{CVPR}, 2016.

\bibitem[Tulyakov et~al.(2018)Tulyakov, Liu, Yang, and Kautz]{tulyakov2018mocogan}
Sergey Tulyakov, Ming-Yu Liu, Xiaodong Yang, and Jan Kautz.
\newblock Mocogan: Decomposing motion and content for video generation.
\newblock In \emph{CVPR}, 2018.

\bibitem[Unterthiner et~al.(2018)Unterthiner, Van~Steenkiste, Kurach, Marinier, Michalski, and Gelly]{unterthiner2018towards}
Thomas Unterthiner, Sjoerd Van~Steenkiste, Karol Kurach, Raphael Marinier, Marcin Michalski, and Sylvain Gelly.
\newblock Towards accurate generative models of video: A new metric \& challenges.
\newblock \emph{arXiv}, 2018.

\bibitem[Wang et~al.(2023{\natexlab{a}})Wang, Li, Lin, Lin, Yang, Zhang, Liu, and Wang]{wang2023disco}
Tan Wang, Linjie Li, Kevin Lin, Chung-Ching Lin, Zhengyuan Yang, Hanwang Zhang, Zicheng Liu, and Lijuan Wang.
\newblock Disco: Disentangled control for referring human dance generation in real world.
\newblock \emph{arXiv}, 2023{\natexlab{a}}.

\bibitem[Wang et~al.(2021)Wang, Mallya, and Liu]{wang2021one}
Ting-Chun Wang, Arun Mallya, and Ming-Yu Liu.
\newblock One-shot free-view neural talking-head synthesis for video conferencing.
\newblock In \emph{CVPR}, 2021.

\bibitem[Wang et~al.(2022)Wang, Yang, Bremond, and Dantcheva]{wang2022latent}
Yaohui Wang, Di Yang, Francois Bremond, and Antitza Dantcheva.
\newblock Latent image animator: Learning to animate images via latent space navigation.
\newblock In \emph{ICLR}, 2022.

\bibitem[Wang et~al.(2023{\natexlab{b}})Wang, Ma, Chen, Dantcheva, Dai, and Qiao]{wang2023leo}
Yaohui Wang, Xin Ma, Xinyuan Chen, Antitza Dantcheva, Bo Dai, and Yu Qiao.
\newblock Leo: Generative latent image animator for human video synthesis.
\newblock \emph{arXiv}, 2023{\natexlab{b}}.

\bibitem[Wang et~al.(2004)Wang, Bovik, Sheikh, and Simoncelli]{wang2004image}
Zhou Wang, Alan~C Bovik, Hamid~R Sheikh, and Eero~P Simoncelli.
\newblock Image quality assessment: from error visibility to structural similarity.
\newblock \emph{IEEE TIP}, 2004.

\bibitem[Wu et~al.(2023)Wu, Ge, Wang, Lei, Gu, Shi, Hsu, Shan, Qie, and Shou]{wu2023tune}
Jay~Zhangjie Wu, Yixiao Ge, Xintao Wang, Stan~Weixian Lei, Yuchao Gu, Yufei Shi, Wynne Hsu, Ying Shan, Xiaohu Qie, and Mike~Zheng Shou.
\newblock Tune-a-video: One-shot tuning of image diffusion models for text-to-video generation.
\newblock In \emph{ICCV}, 2023.

\bibitem[Xiang et~al.(2021)Xiang, Prada, Bagautdinov, Xu, Dong, Wen, Hodgins, and Wu]{xiang2021modeling}
Donglai Xiang, Fabian Prada, Timur Bagautdinov, Weipeng Xu, Yuan Dong, He Wen, Jessica Hodgins, and Chenglei Wu.
\newblock Modeling clothing as a separate layer for an animatable human avatar.
\newblock \emph{ACM TOG}, 2021.

\bibitem[Xu et~al.(2023{\natexlab{a}})Xu, Song, Jiang, Zhang, Shi, Liu, Ma, Feng, and Luo]{xu2023omniavatar}
Hongyi Xu, Guoxian Song, Zihang Jiang, Jianfeng Zhang, Yichun Shi, Jing Liu, Wanchun Ma, Jiashi Feng, and Linjie Luo.
\newblock Omniavatar: Geometry-guided controllable 3d head synthesis.
\newblock In \emph{CVPR}, 2023{\natexlab{a}}.

\bibitem[Xu et~al.(2023{\natexlab{b}})Xu, Zhang, Liew, Feng, and Shou]{xu2023xagen}
Zhongcong Xu, Jianfeng Zhang, Jun~Hao Liew, Jiashi Feng, and Mike~Zheng Shou.
\newblock Xagen: 3d expressive human avatars generation.
\newblock In \emph{NeurIPS}, 2023{\natexlab{b}}.

\bibitem[Ye et~al.(2023)Ye, Zhang, Liu, Han, and Yang]{ye2023ip}
Hu Ye, Jun Zhang, Sibo Liu, Xiao Han, and Wei Yang.
\newblock Ip-adapter: Text compatible image prompt adapter for text-to-image diffusion models.
\newblock \emph{arXiv}, 2023.

\bibitem[Zakharov et~al.(2019)Zakharov, Shysheya, Burkov, and Lempitsky]{zakharov2019few}
Egor Zakharov, Aliaksandra Shysheya, Egor Burkov, and Victor Lempitsky.
\newblock Few-shot adversarial learning of realistic neural talking head models.
\newblock In \emph{CVPR}, 2019.

\bibitem[Zakharov et~al.(2020)Zakharov, Ivakhnenko, Shysheya, and Lempitsky]{zakharov2020fast}
Egor Zakharov, Aleksei Ivakhnenko, Aliaksandra Shysheya, and Victor Lempitsky.
\newblock Fast bi-layer neural synthesis of one-shot realistic head avatars.
\newblock In \emph{ECCV}, 2020.

\bibitem[Zhang et~al.(2023{\natexlab{a}})Zhang, Jiang, Yang, Xu, Shi, Song, Xu, Wang, and Feng]{zhang2023avatargen}
Jianfeng Zhang, Zihang Jiang, Dingdong Yang, Hongyi Xu, Yichun Shi, Guoxian Song, Zhongcong Xu, Xinchao Wang, and Jiashi Feng.
\newblock Avatargen: a 3d generative model for animatable human avatars.
\newblock In \emph{ECCV Workshop}, 2023{\natexlab{a}}.

\bibitem[Zhang et~al.(2023{\natexlab{b}})Zhang, Yan, Xu, Feng, and Liew]{zhang2023magicavatar}
Jianfeng Zhang, Hanshu Yan, Zhongcong Xu, Jiashi Feng, and Jun~Hao Liew.
\newblock Magicavatar: Multimodal avatar generation and animation.
\newblock \emph{arXiv}, 2023{\natexlab{b}}.

\bibitem[Zhang(2023)]{refonly}
Lvmin Zhang.
\newblock Reference-only controlnet.
\newblock \url{https://github.com/Mikubill/sd-webui-controlnet/discussions/1236}, 2023.

\bibitem[Zhang et~al.(2023{\natexlab{c}})Zhang, Rao, and Agrawala]{zhang2023adding}
Lvmin Zhang, Anyi Rao, and Maneesh Agrawala.
\newblock Adding conditional control to text-to-image diffusion models.
\newblock In \emph{ICCV}, 2023{\natexlab{c}}.

\bibitem[Zhang et~al.(2018)Zhang, Isola, Efros, Shechtman, and Wang]{zhang2018unreasonable}
Richard Zhang, Phillip Isola, Alexei~A Efros, Eli Shechtman, and Oliver Wang.
\newblock The unreasonable effectiveness of deep features as a perceptual metric.
\newblock In \emph{CVPR}, 2018.

\bibitem[Zhao and Zhang(2022)]{zhao2022thin}
Jian Zhao and Hui Zhang.
\newblock Thin-plate spline motion model for image animation.
\newblock In \emph{CVPR}, 2022.

\bibitem[Zhou et~al.(2022)Zhou, Wang, Yan, Lv, Zhu, and Feng]{zhou2022magicvideo}
Daquan Zhou, Weimin Wang, Hanshu Yan, Weiwei Lv, Yizhe Zhu, and Jiashi Feng.
\newblock Magicvideo: Efficient video generation with latent diffusion models.
\newblock \emph{arXiv}, 2022.

\end{thebibliography}
